\documentclass[conference]{IEEEtran}
\IEEEoverridecommandlockouts
\usepackage{cite}
\usepackage{amsmath,amssymb,amsfonts}
\usepackage{algorithmic}
\usepackage{graphicx}
\usepackage{textcomp}
\usepackage{multirow}
\usepackage{xcolor}
\usepackage{subcaption}
\usepackage[many]{tcolorbox}
\usepackage{url}
\def\BibTeX{{\rm B\kern-.05em{\sc i\kern-.025em b}\kern-.08em
    T\kern-.1667em\lower.7ex\hbox{E}\kern-.125emX}}
\begin{document}

\title{Fine-Tuning Vision-Language Models for Multimodal Polymer Property Prediction
}

\author{\IEEEauthorblockN{An Vuong}
\IEEEauthorblockA{\textit{Department of EECS} \\
\textit{University of Arkansas}\\
Fayetteville, AR, USA \\
anv@uark.edu}
\and
\IEEEauthorblockN{Minh-Hao Van}
\IEEEauthorblockA{\textit{Department of EECS} \\
\textit{University of Arkansas}\\
Fayetteville, AR, USA \\
haovan@uark.edu}
\and
\IEEEauthorblockN{Prateek Verma}
\IEEEauthorblockA{\textit{Department of EECS} \\
\textit{University of Arkansas}\\
Fayetteville, AR, USA \\
prateek@uark.edu}
\and
\IEEEauthorblockN{Chen Zhao}
\IEEEauthorblockA{\textit{Department of CS} \\
\textit{Baylor University}\\
Waco, TX, USA \\
chen\_zhao@baylor.edu}
\and
\IEEEauthorblockN{Xintao Wu}
\IEEEauthorblockA{\textit{Department of EECS} \\
\textit{University of Arkansas}\\
Fayetteville, AR, USA \\
xintaowu@uark.edu}
}

\newcommand {\pv}[1]{{\color{orange}[\textbf{PV: }#1]}\normalfont}
\newcommand {\hao}[1]{{\color{blue}[\textbf{Hao: }#1]}\normalfont}
\newcommand {\xw}[1]{{\color{red}[\textbf{XW: }#1]}\normalfont}

\maketitle

\begin{abstract}
Vision-Language Models (VLMs) have shown strong performance in tasks like visual question answering and multimodal text generation, but their effectiveness in scientific domains such as materials science remains limited. While some machine learning methods have addressed specific challenges in this field, there is still a lack of foundation models designed for broad tasks like polymer property prediction using multimodal data. In this work, we present a multimodal polymer dataset to fine-tune VLMs through instruction-tuning pairs and assess the impact of multimodality on prediction performance. Our fine-tuned models, using LoRA, outperform unimodal and baseline approaches, demonstrating the benefits of multimodal learning. Additionally, this approach reduces the need to train separate models for different properties, lowering deployment and maintenance costs.

\end{abstract}


\section{Introduction}
Vision-Language Models (VLMs) have demonstrated exceptional capabilities in visio-linguistic tasks such as visual question answering (VQA), multimodal information extraction, and complex multimodal reasoning. A typical VLM consists of three main components: a vision encoder that extracts visual embeddings from input images, a large language model (LLM) that generates output tokens, and a multimodal projector that maps visual embeddings into a textual space processable by the language model.
% Together, these components enable VLMs to process multimodal information and generate text conditioned on both images and task-specific instructions. 
While both LLMs and VLMs have proven effective for reasoning tasks in general knowledge domains, applying them to specialized scientific areas, such as materials science, remains an open challenge.

% Trained on vast corpora of human knowledge, LLMs can handle both general queries and domain-specific scientific tasks requiring deep reasoning. When paired with vision encoders, they extend into Vision–Language Models (VLMs) that jointly process text and images. While successful in natural image domains, VLMs face challenges in scientific contexts where visual structures are complex and relevant training data are scarce. 
In polymer research, SMILES (Simplified Molecular-Input Line-Entry System) \cite{weininger1988smiles}, which is a text-based representation of molecular structures, has been extended by introducing asterisks (*) to mark the repeating units, referred to as polymer SMILES (P-SMILES) \cite{kuenneth2023polybert}. Recent studies have explored fine-tuning LLMs for property prediction \cite{gupta2025benchmarking} and combining LLM embeddings with conformational features \cite{zhang2025multimodal}. Yet, these approaches are either unimodal or rely on separate regressors and therefore lack unified multimodal alignment. Earlier machine learning efforts also contributed to property prediction but often required training separate models for each task, creating fragmented pipelines. To overcome these limitations, we introduce a multimodal polymer dataset tailored for fine-tuning and evaluating VLMs, and investigate their ability to directly predict polymer properties. Specifically: (1) Our multimodal dataset is built from computational and experimental sources. Each sample contains a canonical P-SMILES, a 2D structure image, molecular descriptors, and property labels. The dataset is structured as VQA pairs, enabling VLMs to predict properties from images, P-SMILES, and molecular descriptors with improved reasoning. (2) We evaluate multimodal VLMs on polymer property prediction by fine-tuning Llama \cite{meta2024llama} and Qwen \cite{bai2023qwen} vision-language and text-only models. Using instruction tuning with LoRA \cite{hu2022lora}, the fine-tuned model achieves performance on par with machine learning and deep learning methods that typically require separate models for each property.

\section{Related Works}
Beyond traditional deep learning models designed for classification, vision-language models have recently gained significant attention due to their broad range of applications. CLIP \cite{radford2021learning} introduced a framework that links image and text data, enabling models to learn visual concepts for classification tasks through natural language supervision. Building on this idea, models such as OpenAI GPT-4 \cite{OpenAI2023GPT4TR}, Google Gemini \cite{team2023gemini}, Flamingo \cite{alayrac2022flamingo}, OpenFlamingo \cite{awadalla2023openflamingo}, LLama-Vision \cite{meta2024llama} and LLaVA \cite{liu2023visual} have demonstrated strong performance on vision-language tasks, including visual question answering and human-like dialogue. Alongside these foundation models, researchers have explored extending large-scale architectures to specialized domains, such as LLaVA-Med \cite{li2023llava}, Llama-SciTune \cite{horawalavithana2023scitune}, and Vision-BioLLM \cite{alshibli2025vision}.

There have been numerous efforts to integrate AI into materials science research, particularly through the development of deep learning and foundation models for materials discovery and property prediction. This is especially relevant for polymers, where the cost of computational simulations or experimental measurements is often prohibitively high. Huan et al. \cite{huan2016polymer} introduced a dataset of polymer properties, which laid the foundation for the Polymer Genome platform \cite{kim2018polymer} designed to efficiently predict and retrieve polymer properties. Building on this, Doan et al. \cite{doan2020machine} applied machine learning approaches trained on Polymer Genome data for property prediction. More recently, BERT-based models \cite{devlin2019bert} have been adapted for polymers: Kuenneth et al. \cite{kuenneth2023polybert} developed PolyBERT, a large-scale representation model trained on millions of polymers, whose embeddings can serve as inputs for property predictors. Similarly, Wang et al. \cite{wang2024predicting} proposed a Transformer-based architecture capable of extracting both 1D representations from P-SMILES and 3D representations from molecular conformations to perform multitask learning, including P-SMILES reconstruction, 3D coordinate generation, and cross-modal fusion. In line with recent trends in deep learning, large language models (LLMs) have also been explored for polymer property prediction \cite{gupta2025benchmarking,zhang2025multimodal}. Specifically, Gupta et al. \cite{gupta2025benchmarking} fine-tuned and evaluated text-only LLMs on P-SMILES inputs, while Zhang et al. \cite{zhang2025multimodal} combined multimodal embeddings, such as LLM-derived representations from P-SMILES and Uni-Mol \cite{zhou2023uni} embeddings from polymer structures, and then train multilayer perceptrons for property prediction. Despite these advances, a unified multimodal Vision-Language Model capable of directly predicting properties from multimodal inputs such as images, text, and molecular descriptors remains lacking.

\section{Preliminary}

\subsection{Large-Language Models}
LLMs have achieved remarkable success in bridging artificial intelligence (AI) with real-world applications. Built upon Transformer decoder-only layers \cite{vaswani2017attention}, LLMs generate answers and content by predicting the next token conditioned on a user input prompt. Trained on large corpora of human knowledge, LLMs exhibit impressive capabilities, including generating contextually coherent responses, performing reasoning on complex tasks, enabling zero-shot and few-shot learning, supporting code generation, and handling multiple downstream tasks—capabilities that traditional deep learning architectures such as CNNs or LSTMs cannot achieve. Recent state-of-the-art LLMs, such as Llama \cite{meta2024llama}, Qwen \cite{bai2023qwen}, Claude \cite{anthropic2024claude3}, ChatGPT \cite{OpenAI2023GPT4TR}, and Google Gemini \cite{team2023gemini}, have demonstrated outstanding performance. Moreover, many of these LLMs have been extended into Vision–Language Models (VLMs), enabling them to process both textual and visual inputs.

However, the computational cost of training or fine-tuning large models remains prohibitively high, limiting accessibility for practitioners with modest resources. To address this challenge, several efficient techniques have been proposed. Without the need for fine-tuning, zero-shot and few-shot approaches \cite{kojima2022large,brown2020language} adapt LLMs to downstream tasks by designing effective prompts or providing task-specific demonstrations. Chain-of-Thought prompting \cite{wei2022chain} further enhances reasoning by structuring tasks into multi-step instructions, where intermediate outputs serve as inputs for subsequent steps.

For scenarios with moderate resources, parameter-efficient fine-tuning offers a practical alternative to full-model training. Soft prompting methods, such as P-tuning and prefix-tuning, adapt LLMs to new tasks by prepending learnable tokens to the input sequence. During fine-tuning, the model parameters are frozen, and only these task-specific tokens are optimized. Low-Rank Adaptation (LoRA) \cite{hu2022lora} is another widely adopted technique for efficient fine-tuning, particularly in VLMs. LoRA decomposes large weight matrices in attention layers into low-rank components, substantially reducing the number of trainable parameters. Since its introduction, several variants have been proposed to further improve efficiency and robustness, including QLoRA \cite{dettmers2023qlora}, DoRA \cite{liu2024dora}, AdaLoRA \cite{zhang2023adalora}, and VeRA \cite{kopiczko2023vera}.

\subsection{Vision-Language Models}
LLMs have been extended into VLMs with the capability to process and understand visual inputs. To enable this, a vision encoder extracts patch embeddings from images, which are then projected into the textual embedding space via a multimodal projector (typically a linear layer). Common choices for vision encoders include CLIP \cite{radford2021learning} and SigLIP \cite{zhai2023sigmoid}. During decoding, visual context can be integrated into the language model in two standard ways: (i) inserting cross-modal attention layers into the LM architecture, or (ii) converting visual embeddings into contextual “visual tokens” that are fed directly to the LM. In the first design, a cross-modal attention block functions as a cross-attention layer in which the keys and values are derived from visual embeddings. LM layers interleave self-attention with cross-attention, allowing the model to attend to visual information throughout generation and reasoning. This approach underlies the Llama-3.2-Vision family \cite{grattafiori2024llama} and the Flamingo/OpenFlamingo family \cite{alayrac2022flamingo,awadalla2023openflamingo}. In the latter design, visual embeddings are treated as tokens inserted at specific positions within the text sequence; the resulting multimodal sequence is then processed by a decoder-only LM using self-attention alone. Here, visual tokens serve as additional context, enabling joint attention over image and text. State-of-the-art VLMs such as Qwen-2.5-VL \cite{bai2025qwen2} and LLaVA-1.5 \cite{liu2023visual} adopt this strategy.

\section{Fine-tuning Vision-Language Models for Multimodal Polymer Property Prediction}

\subsection{Multimodal Polymer Dataset}
\begin{figure*}[ht]  
  \centering
  \includegraphics[width=1\textwidth]{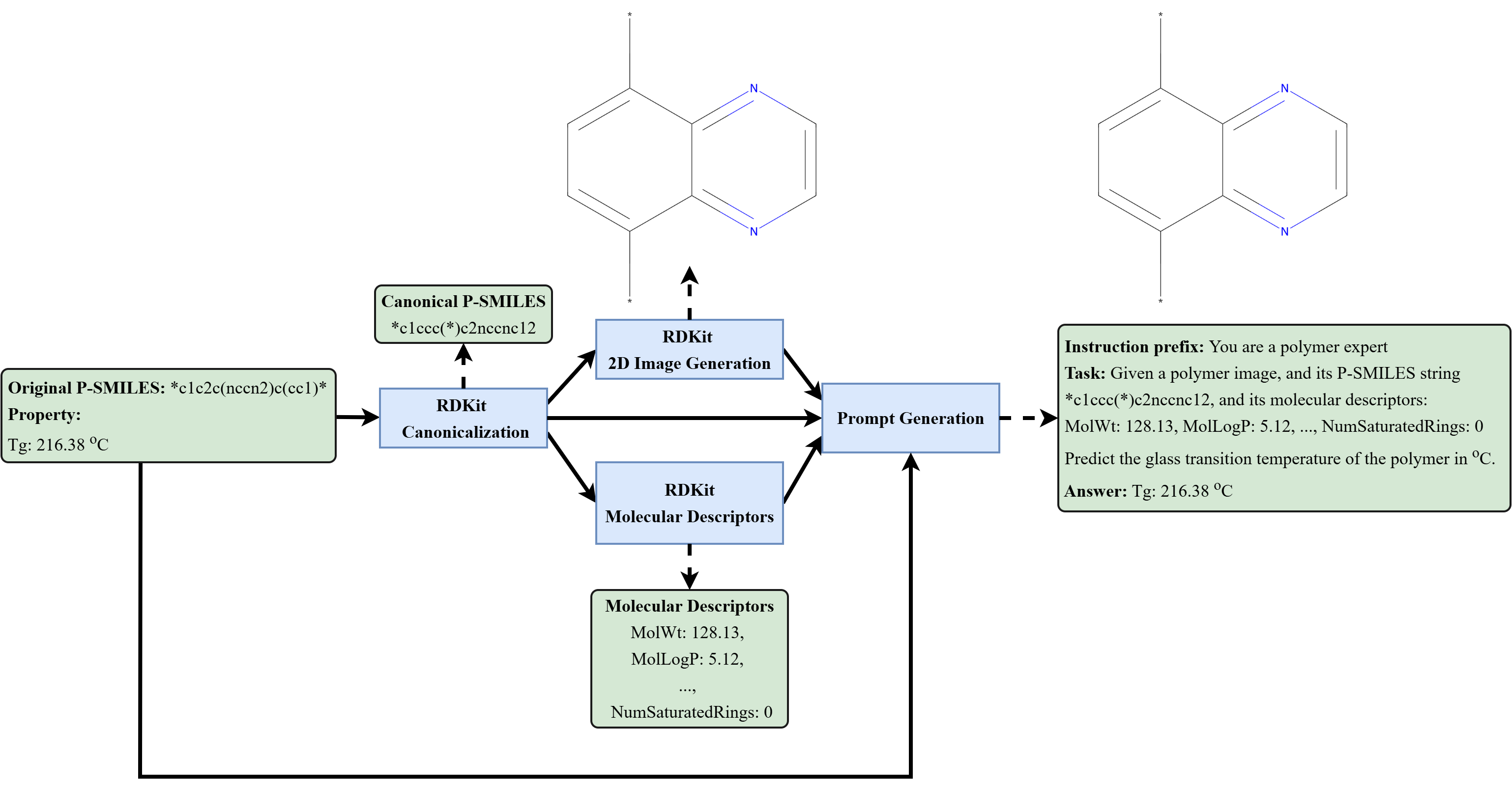} %
  \caption{Pipeline for multimodal prompt generation.}
  \label{fig:pipeline}
\end{figure*}

In this section, we present our framework for generating a Multimodal Polymer Dataset including 3 main stages: (1) Data Collection and Preprocessing, (2) Multimodal Feature Generation, (3) Instruction-Tuning Dataset Generation. Figure \ref{fig:pipeline} illustrates our pipeline for multimodal prompt generation.

\subsubsection{Data Collection and Preprocessing}

We collect the data from the Kaggle Open Polymer Prediction 2025 \cite{liu2025kaggle_challenge}, including 7,973 P-SMILES with five properties: glass transition temperature (Tg), fractional free volume (FFV), thermal conductivity (Tc), density, and radius of gyration (Rg). Besides the main dataset, there are four supplementary datasets provided in this challenge, but we only use three of them: the first with 874 P-SMILES and Tc values, the third with 46 P-SMILES and Tg values, and the fourth with 862 P-SMILES and FFV values. The second supplementary dataset, having 7,174 P-SMILES without any property values, is not used in our study.

\noindent\textbf{Data preprocessing.} To combine the supplementary datasets with the main Kaggle dataset, we first canonicalize P-SMILES strings in all datasets to obtain a unique representation for each polymer. We then remove duplicates from the supplementary datasets by checking against the main dataset. A polymer is considered a duplicate if it has the same canonical P-SMILES and the same property values as a polymer already included in the main dataset. Following this process, we sequentially merge the main dataset with the first, third and fourth datasets. Duplicates are removed if found.

As a result, the final dataset consists of 8,963 P-SMILES, each with a varying number of ground-truth properties. Table~\ref{tab:data_stat} summarizes the data statistics, including missing values across the five properties. Finally, we split the dataset into training and testing sets in a 90/10 ratio based on canonical P-SMILES, ensuring no polymer appears in both the training and test sets. The split results in 7,950 polymers for training and 1,013 for testing. 

\begin{table}[ht]
    \centering
    \caption{Number of samples for each property in Kaggle dataset.}
    \begin{tabular}{|l|ccccc|}
        \hline
         & Tg & FFV & Tc & Density & Rg  \\
        \hline
        Missing count & 8400 & 1071 & 8106 & 8350  & 8349  \\
        Missing ratio & 93.72\% & 11.95\% &  90.44\% & 93.16\% & 93.14\%  \\
        \hline
    \end{tabular}
    \label{tab:data_stat}
\end{table}

To evaluate model's performance on unknown dataset, we use Glass Transition Temperature dataset (GTT), which contains 662 P-SMILES with Tg values introduced by Choi et al. \cite{choi2024automated}. We apply the same preprocessing steps to GTT and then compare it with the final dataset to filter out duplicates, resulting in 563 duplicates being removed and 99 polymers with Tg values retained for model testing. Similarly, we also conduct evaluation on RadonPy dataset \cite{hayashi2022radonpy}, including 1,077 polymers with three properties (Tc, Density, Rg). After applying preprocessing steps and filtering out duplicates with Kaggle dataset, the final version used in our experiment consists of 292 P-SMILES. Among these, 4 P-SMILES lack Tc values, while the others retain all three properties.

\subsubsection{Multimodal Features Generation}

Fore both the training and testing sets, we generate a 2D image of each polymer from its canonical P-SMILES using RDKit \cite{rdkit}, at a resolution of 1120 × 1120 pixels. We also compute 217 molecular descriptors from each P-SMILES with the RDKit Python package. Of these, 17 descriptors were selected by domain experts as meaningful features for models to predict the five target properties. Table \ref{tab:descriptors} shows the 17 molecular descriptors used in our dataset (descriptor descriptions are adopted from \cite{epa2007descriptors,datagrok_molecular_descriptors}): 

\begin{table}[ht]
    \centering
    \caption{Molecular descriptors used in multimodal polymer dataset}
    \begin{tabular}{|l|l|}
        \hline
        Descriptor & Description \\
        \hline
       MolWt  &  Molecular weight \\
       MolLogP  & Octanol–water partition coefficient (logP) \\
       BalabanJ & Balaban’s J topological index \\
       Chi0 & Zero-order molecular connectivity index \\
       Chi1 & First-order molecular connectivity index \\
       HallKierAlpha & Hall–Kier alpha parameter \\
       LabuteASA & Labute’s Approximate Surface Area \\
       TPSA & The polar surface area of a molecule based \\
       & upon fragments \\
       FractionCSP3 & The fraction of C atoms that are SP3\\
       & hybridized \\
       HeavyAtomCount & The number of heavy atoms \\
       NHOHCount & The number of NHs or OHs \\
       NOCount & The number of nitrogens and oxygens \\
       NumAliphaticRings & The number of aliphatic rings \\
       NumAmideBonds & The number of amide bonds \\
       NumAromaticRings & The number of aromatic rings \\
       NumRotatableBonds & The number of rotatable bonds \\
       NumSaturatedRings & The number of saturated rings \\
       \hline
    \end{tabular}
    \label{tab:descriptors}
\end{table}

\subsubsection{Instruction-tuning Dataset Generation}

We construct an instruction-tuning dataset to predict one property type at a time. For polymers with multiple ground-truth properties, each data sample is decomposed into separate instruction-tuning samples that share the same canonical P-SMILES representation but are assigned different prompts, each requesting the prediction of a single property type with an available ground-truth value. After this decomposition, the training set expands to 9,097 samples and the testing set to 1,442 samples. Each sample is then structured as a question–answer pair, where the question corresponds to the prompt and the answer contains the ground-truth property value. The prompt is generated by randomly combining one of 20 instruction prefixes with one of 20 prediction templates, into which the canonical P-SMILES, the 2D image produced with RDKit, the 17 selected descriptors, and the specific property type to be predicted are inserted. The answer is standardized to a consistent format for each property type. Figure \ref{fig:prompt} shows an example of our instruction-tuning sample.

\begin{figure}[ht]
    \centering
    \begin{tcolorbox}[colback={rgb,255:red,213;green,232;blue,212},colframe=green!50!blue!100,left=2pt, right=2pt, top=2pt, bottom=2pt] % title=Definition-guided Few-shot Prompt
        \textbf{Instruction prefix:} You are a polymer expert.\\
        \textbf{Task:} Given a polymer image \texttt{<image>}, its P-SMILES string \texttt{<P-SMILES>}, and its molecular descriptors: \\
        \texttt{<descriptor-1:value>},
        \texttt{<descriptor-2:value>}, \\
        \dots, \\
        \texttt{<descriptor-17:value>}. \\
        Predict the \texttt{<property type>} of the polymer in \texttt{<unit>}.\\
        \textbf{Answer:} \texttt{<property type>:<property value>} \texttt{<unit>}
    \end{tcolorbox}  
    \caption{An example of our instruction-tuning sample.}
    \label{fig:prompt}
\end{figure}

\subsection{Fine-tuning VLMs with LoRA}
\begin{figure*}[ht]
    \centering
    \includegraphics[width=0.9\linewidth]{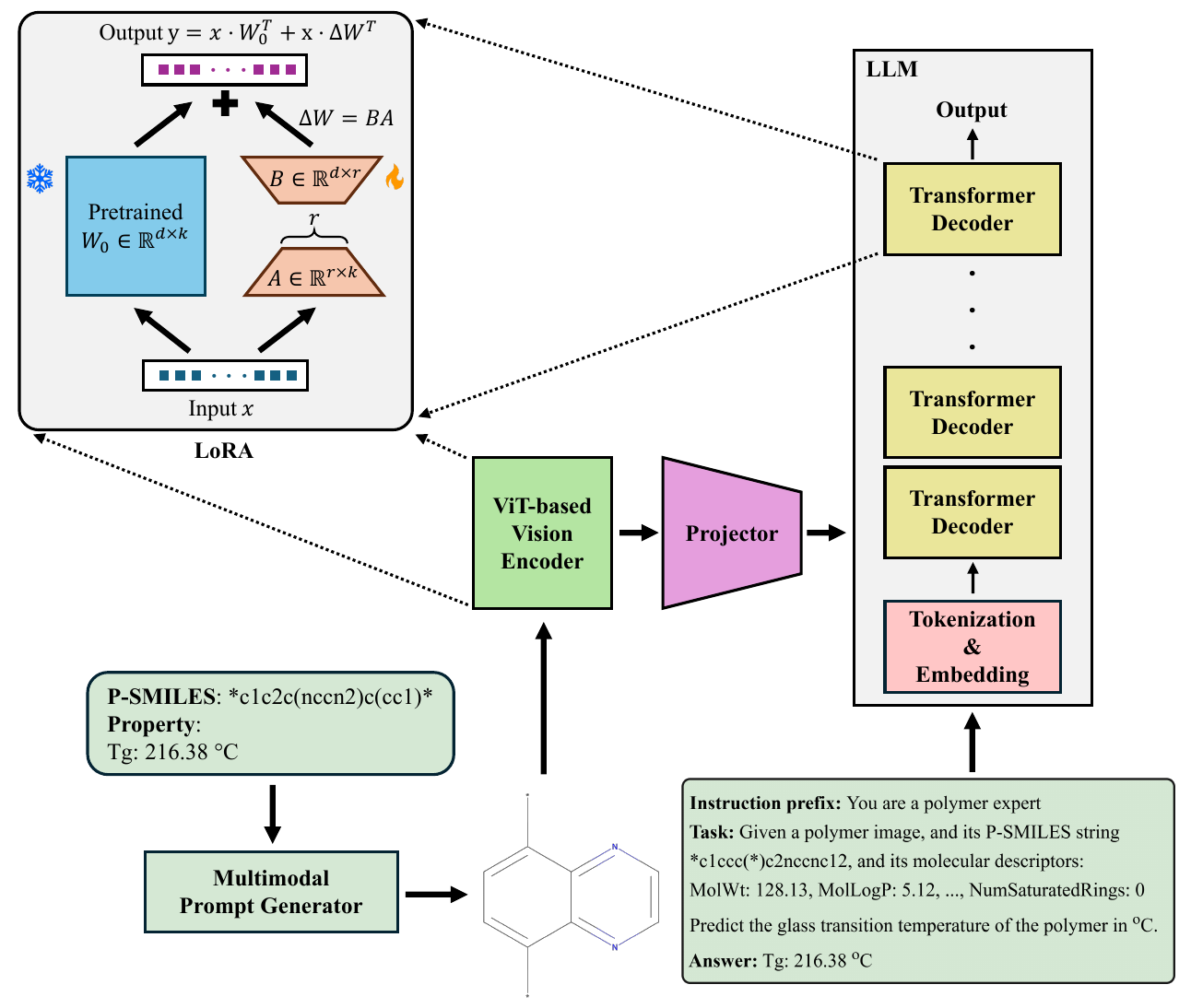}
    \caption{An illustrative figure of fine-tuning VLMs with LoRA. Parameters of Transformer blocks in LLM and Vision Encoder are decomposed into two low-rank matrices that will be updated.}
    \label{fig:lora_finetune}
\end{figure*}

As discussed above, Low-Rank Adaptation (LoRA) was introduced to reduce the cost of fine-tuning LLMs and VLMs by significantly decreasing the number of trainable parameters. Instead of updating the full set of parameters in large weight matrices, LoRA optimizes two low-rank matrices that approximate the parameter updates in attention layers. Specifically, let $W_0 \in \mathbb{R}^{d \times k}$ denote a pretrained model weight matrix where $k$ and $d$ are input and output dimensions of the fine-tuned layer. In the full fine-tuning setting, all $d \times k$ parameters of $W_0$ are updated. In contrast, LoRA keeps $W_0$ frozen and learns additional low-rank matrices:

\begin{align*}
W &= W_0 + \Delta W, \\
\Delta W &= BA,
\end{align*}
where $A \in \mathbb{R}^{r \times k}$, $B \in \mathbb{R}^{d \times r}$, and $r \ll \min(d,k)$ denotes the rank constraint. Thus, instead of training $d \times k$ parameters, LoRA requires updating only $r(d+k)$ parameters. For example, the query projection matrix $W_q$ in the Llama-3.1-8B model has dimensions $4096 \times 4096$, amounting to approximately 16 million parameters. With LoRA using rank $r=8$, the number of trainable parameters is reduced to $8(4096+4096) \approx 65{,}000$, yielding a 240-fold reduction. Figure~\ref{fig:lora_finetune} illustrates how LoRA can be applied to fine-tune VLMs for multimodal polymer property prediction.

\section{Experiments}

\subsection{Models}
\noindent\textbf{Our models.} We fine-tune the following LLMs and VLMs: Llama-3.2-11B-Vision-Instruct (LVision), Llama-3.1-8B-Instruct (LText), Qwen2.5-VL-7B-Instruct (QVision), and Qwen2.5-7B-Instruct (QText). The LoRA \cite{hu2022lora} technique is adopted to fine-tune models with $\text{rank}=16$ and $\alpha=16$. All models are fine-tuned for 12 epochs with a learning rate of $0.0001$, a batch size of 8, a gradient accumulation step of 4 and weight decay of 0.01.

\noindent\textbf{Baselines.}
To validate our approach, we compare the fine-tuned LVision against three baseline groups: (1) ML models using molecular descriptors, (2) ML models using PolyBERT-derived representations \cite{kuenneth2023polybert}, and (3) Pretrained LLMs and VLMs. For the descriptor-based ML group (+RDKit), we train Multi-layer Perceptron (MLP), Support Vector Regressor (SVR), Random Forest (RF), and Linear Regression (LinR) models using 17 molecular descriptors. For the PolyBERT-based ML group (+PolyBERT), we extract polymer representations from a pretrained PolyBERT and train the same ML models. For the pretrained LLM and VLM baselines, we evaluate the original (non-fine-tuned) LText, LVision, QText, and QVision. Each property is predicted using a separate model, resulting in five models for five properties.

\noindent\textbf{Running time.} We perform fine-tuning and evaluation on an H200 GPU with 141 GB of RAM. Fine-tuning LVision takes about 21 hours for 12 epochs with $1120\times1120$ images, while fine-tuning LText takes about one hour under the same settings.

\subsection{Metrics}
In our experiments, we report the Mean Absolute Error (MAE) and Mean Absolute Percentage Error (MAPE), since the task involves predicting continuous property values. To evaluate overall model performance across multiple properties, we also report the Weighted Mean Absolute Error (wMAE), as introduced in the Kaggle challenge \cite{liu2025kaggle_challenge}. 

\textbf{MAE.}
The Mean Absolute Error (MAE) measures the average absolute difference between the predicted and ground truth values of a single property type:

\begin{equation*}
    \text{MAE} = \frac{1}{n} \sum_{i=1}^{n} \big| \hat{y}_i - y_i \big|,
\end{equation*}
where $n$ denotes the number of available ground-truth values for a property type under evaluation, $y_i$ and $\hat{y}_i$ represent the ground-truth and predicted values of the $i$-th polymer, respectively.

\textbf{MAPE.}
The Mean Absolute Percentage Error (MAPE) measures the average absolute percentage error between the predicted and ground truth values of a single property type:
\begin{equation*}
    \text{MAPE} = \frac{100}{n} \sum_{i=1}^{n} \frac{\left| \hat{y}_i - y_i \right|}{|y_i|}
\end{equation*}
% where $n$ denotes the number of available ground-truth values for the property type under evaluation, $y_i$ and $\hat{y}_i$ represent the ground-truth and predicted values of the $i$-th polymer, respectively.

\textbf{wMAE.}
Weighted Mean Absolute Error (wMAE) is the evaluation metric used in the Kaggle contest \cite{liu2025kaggle_challenge} 
to evaluate the overall prediction performance across five properties:

\begin{align*}
\text{wMAE} &= \frac{1}{n} \sum_{i=1}^n \sum_{k=1}^{\mathcal{I}_i} 
w_k \cdot \left| \hat{y}_i^k - y_i^k \right|, \\
w_k &= \left(\frac{1}{r_k}\right) \cdot 
\left( \frac{K \cdot \sqrt{1/n_k}}{\sum_{j=1}^K \sqrt{1/n_j}} \right),
\label{eq:wmae}
\end{align*}
where $n$ is the number of polymer being evaluated, $\mathcal{I}_i$ denotes the set of property types of the $i$-th polymer, and $\hat{y}_i^k$ and $y_i^k$ are the predicted and ground-truth values of the property $k$  of polymer $i$-th, respectively. Moreover, $w_k$ is reweighting factor for each property where $n_k$ denotes the number of samples having $k$-th property, $K$ is the number of property types, and $r_k = \max(y^k) - \min(y^k)$ represents the estimated value range of the $k$-th property based on the test data.

\subsection{Results and Discussion}

\begin{table*}[ht]
    \centering
    \caption{Comparison of fine-tuned Llama and Qwen models with baseline approaches for predicting five polymer properties on Kaggle test dataset. Results are reported as mean and standard deviation of MAE, with the best performance for each column highlighted in bold.}
    \begin{tabular}{|l|cccccc|}
        \hline
        \multirow{3}{*}{Model} & \multicolumn{6}{c|}{Kaggle} \\
        \cline{2-7}
        & Tg$\downarrow$ & FFV$\downarrow$ & Tc$\downarrow$ & Density$\downarrow$ & Rg$\downarrow$ & wMAE$\downarrow$ \\
        & & \small{$\times10^{-2}$} & \small{$\times10^{-2}$} & \small{$\times10^{-2}$} &  & \small{$\times10^{-2}$} \\
        \hline
        MLP + RDKit & $\mathbf{46.5_{3.7}}$ & $6.6_{1.5}$ & $14.6_{4.9}$ & $17.9_{4.5}$ & $3.7_{0.2}$ & $13.1_{1.3}$ \\
        SVR + RDKit & $92.8_{0.0}$ & $2.3_{0.0}$ & $3.8_{0.0}$ & $5.9_{0.0}$ & $3.3_{0.0}$ & $7.4_{0.0}$ \\
        RF + RDKit & $55.9_{0.2}$ & $1.2_{0.0}$ & $3.9_{0.0}$ & $7.7_{0.0}$ & $3.1_{0.0}$ & $6.1_{0.0}$ \\
        LinR + RDKit & $54.9_{0.0}$ & $1.4_{0.0}$ & $3.7_{0.0}$ & $6.4_{0.0}$ & $3.1_{0.0}$ & $6.0_{0.0}$ \\
        \hline
        MLP + PolyBERT & $67.9_{2.0}$ & $1.9_{0.3}$ & $5.4_{0.7}$ & $7.8_{0.6}$ & $\mathbf{1.8_{0.0}}$ & $5.9_{0.2}$  \\
        SVR + PolyBERT & $75.2_{0.0}$ & $1.9_{0.0}$ & $3.6_{0.0}$ & $4.7_{0.0}$ & $\mathbf{1.8_{0.0}}$ & $5.3_{0.0}$ \\
        RF + PolyBERT & $64.5_{0.3}$ & $1.2_{0.0}$ & $\mathbf{3.1_{0.0}}$ & $6.9_{0.0}$ & $2.5_{0.0}$ & $5.5_{0.0}$ \\
        LinR + PolyBERT & $168.8_{0.0}$ & $1.0_{0.0}$ & $13.1_{0.0}$ & $6.7_{0.0}$ & $6.1_{0.0}$ & $11.4_{0.0}$ \\
        \hline
        Original LText & $100.2_{1.3}$ & $4.9_{0.2}$ & $7.9_{0.2}$ & $14.3_{0.5}$ & $5.9_{0.3}$ & $13.3_{0.3}$  \\
        Original LVision & $90.3_{1.8}$ & $4.4_{0.0}$ & $8.1_{0.2}$ & $9.6_{0.2}$ & $7.2_{0.4}$ & $13.0_{0.2}$ \\
        Original QText & $108.3_{5.9}$ & $9.9_{0.2}$ & $9.2_{0.3}$ & $26.4_{0.6}$ & $9.6_{0.1}$ & $22.0_{0.3}$ \\
        Original QVision & $97.9_{1.4}$ & $5.3_{0.1}$ & $10.9_{0.1}$ & $8.4_{0.3}$ & $6.9_{0.3}$ & $13.6_{0.2}$ \\
        \hline
        Fine-tuned LText & $60.0_{3.5}$ & $1.0_{0.0}$ & $5.7_{0.5}$ & $4.0_{0.1}$ & $2.8_{0.1}$ & $5.4_{0.1}$  \\
        Fine-tuned LVision & $58.0_{2.7}$ & $1.0_{0.0}$ & ${3.6_{0.1}}$ & $\mathbf{3.3_{0.1}}$ & $2.3_{0.0}$ & $\mathbf{4.5_{0.0}}$  \\
        Fine-tuned QText & $64.8_{1.9}$ & $\mathbf{0.9_{0.0}}$ & $3.3_{0.1}$ & $5.6_{0.3}$ & $2.2_{0.0}$ & $4.9_{0.1}$\\
        Fine-tuned QVision & $68.8_{3.4}$ & $1.0_{0.0}$ & $3.9_{0.1}$ & $3.6_{0.2}$ & $2.7_{0.1}$ & $5.2_{0.2}$ \\
        \hline
    \end{tabular}
    \label{tab:kaggle_mae_results}
\end{table*}

\begin{table*}[ht]
    \centering
    \caption{Comparison of fine-tuned Llama and Qwen models with baseline approaches on unseen datasets (GTT and RadonPy). Results are reported as mean and standard deviation of MAE, with the best performance for each column highlighted in bold.}
    \begin{tabular}{|l|c|cccc|}
        \hline
        \multirow{3}{*}{Model} & GTT & \multicolumn{4}{c|}{RadonPy}\\
        \cline{2-6}
        & Tg$\downarrow$ & Tc$\downarrow$ & Density$\downarrow$  & Rg$\downarrow$ & wMAE$\downarrow$ \\
        & & \small{$\times10^{-2}$} & \small{$\times10^{-2}$} &  & \small{$\times10^{-2}$} \\
        \hline
        MLP + RDKit & $\mathbf{61.4_{7.7}}$ & $17.4_{8.8}$ & $23.3_{2.1}$ & $7.9_{0.5}$ & $73.7_{6.9}$ \\
        SVR + RDKit &  $97.9_{0.0}$ & $5.2_{0.0}$ & $8.8_{0.0}$ & $7.0_{0.0}$ & $46.6_{0.0}$ \\
        RF + RDKit & $67.3_{0.1}$ & $5.5_{0.0}$ & $12.2_{0.0}$ & $6.6_{0.0}$ & $48.6_{0.1}$ \\
        LinR + RDKit & $62.2_{0.0}$ & $5.7_{0.0}$ & $10.6_{0.0}$ & $6.3_{0.0}$ & $45.6_{0.0}$ \\
        \hline
        MLP + PolyBERT &  $67.5_{2.7}$ & $10.0_{1.6}$ & $13.3_{0.8}$ & $\mathbf{4.9_{0.2}}$ & $44.0_{1.5}$ \\
        SVR + PolyBERT & $77.9_{0.0}$ & $\mathbf{4.8_{0.0}}$ & $9.3_{0.0}$ & $5.7_{0.0}$ & $\mathbf{40.6_{0.0}}$ \\
        RF + PolyBERT & $67.8_{0.2}$ & $5.2_{0.0}$ & $10.2_{0.0}$ & $6.8_{0.0}$ & $47.2_{0.0}$ \\
        LinR + PolyBERT & $161.1_{0.0}$ & $26.2_{0.0}$ & $11.8_{0.0}$ & $9.8_{0.0}$ & $77.2_{0.0}$ \\
        \hline
        Original LText & $104.7_{1.8}$ & $8.0_{0.2}$ & $10.5_{0.1}$ & $10.4_{0.1}$ & $66.8_{1.2}$ \\
        Original LVision & $97.4_{1.3}$ & $5.9_{0.1}$ & $8.6_{0.1}$ & $10.0_{0.2}$ & $61.7_{1.2}$ \\
        Original QText & $103.3_{3.9}$ & $7.4_{0.4}$ & $19.4_{0.6}$ & $16.9_{0.1}$ & $107.5_{1.1}$ \\
        Original QVision & $85.9_{3.5}$ & $11.8_{0.2}$ & $14.0_{0.2}$ & $13.3_{0.1}$ & $87.0_{0.9}$ \\
        \hline
        Fine-tuned LText & $70.7_{3.3}$ & $6.0_{0.3}$ & $11.2_{0.1}$ & $6.1_{0.2}$ & $45.0_{1.1}$  \\
        Fine-tuned LVision & $67.7_{2.5}$ & $6.3_{0.1}$ & $\mathbf{7.1_{0.2}}$ & $6.5_{0.1}$ & $43.1_{0.4}$\\
        Fine-tuned QText & $81.4_{5.6}$ & $6.7_{0.0}$ & $\mathbf{7.1_{0.2}}$ & $6.2_{0.1}$ & $42.0_{0.6}$ \\
        Fine-tuned QVision & $65.9_{3.6}$ & $6.2_{0.2}$ & $8.2_{0.1}$ & $6.0_{0.1}$ & $41.5_{0.4}$ \\
        \hline
    \end{tabular}
    \label{tab:unseen_mae_results}
\end{table*}

\begin{table*}[ht]
    \centering
    \caption{Comparison of fine-tuned Llama and Qwen models with baseline approaches for predicting five polymer properties. Results are reported as mean and standard deviation of MAPE, with the best performance for each column highlighted in bold.}
    \begin{tabular}{|l|ccccc|c|ccc|}
        \hline
        \multirow{2}{*}{Model} & \multicolumn{5}{c|}{Kaggle} & GTT & \multicolumn{3}{c|}{RadonPy} \\
        \cline{2-10}
         & Tg$\downarrow$ & FFV$\downarrow$ & Tc$\downarrow$ & Density$\downarrow$ & Rg$\downarrow$ & Tg$\downarrow$ & Tc$\downarrow$ & Density$\downarrow$  & Rg$\downarrow$ \\
        \hline
        MLP + RDKit & $118.2_{8.3}$ & $18.2_{3.9}$ & $70.1_{23.3}$ & $17.4_{4.7}$ & $23.0_{1.6}$ & $\mathbf{75.4_{4.9}}$ & $73.3_{38.0}$ & $19.1_{1.9}$ & $27.3_{1.7}$ \\
        SVR + RDKit & $190.5_{0.0}$ & $6.3_{0.0}$ & $17.1_{0.0}$ & $5.7_{0.0}$ & $18.8_{0.0}$ & $122.4_{0.0}$ & $\mathbf{17.7_{0.0}}$ & $7.2_{0.0}$ & $23.4_{0.0}$ \\
        RF + RDKit & $116.3_{0.5}$ & $3.1_{0.0}$ & $18.6_{0.0}$ & $7.3_{0.0}$ & $18.9_{0.0}$ & $86.5_{0.4}$ & $20.8_{0.0}$ & $9.7_{0.0}$ & $21.4_{0.0}$ \\
        LinR + RDKit & $147.3_{0.0}$ & $3.9_{0.0}$ & $16.9_{0.0}$ & $6.2_{0.0}$ & $18.6_{0.0}$ & $87.2_{0.0}$ & $20.0_{0.0}$ & $8.6_{0.0}$ & $21.3_{0.0}$ \\
        \hline
        MLP + PolyBERT & $133.1_{12.6}$ & $5.2_{0.8}$ & $25.5_{3.6}$ & $7.6_{0.7}$ & $\mathbf{9.7_{0.1}}$ & $113.1_{21.8}$ & $40.0_{7.1}$ & $11.1_{0.6}$ & $\mathbf{16.7_{0.8}}$ \\
        SVR + PolyBERT & $152.1_{0.0}$ & $5.1_{0.0}$ & $16.9_{0.0}$ & $4.5_{0.0}$ & $9.9_{0.0}$ & $92.2_{0.0}$ & $\mathbf{17.7_{0.0}}$ & $7.5_{0.0}$ & $18.8_{0.0}$ \\
        RF + PolyBERT & $142.0_{1.2}$ & $3.1_{0.0}$ & $14.7_{0.0}$ & $6.6_{0.0}$ & $14.9_{0.0}$ & $97.2_{0.3}$ & $19.2_{0.0}$ & $8.2_{0.0}$ & $22.2_{0.0}$ \\
        LinR + PolyBERT & $543.2_{0.0}$ & $2.7_{0.0}$ & $64.1_{0.0}$ & $6.9_{0.0}$ & $38.7_{0.0}$ & $337.2_{0.0}$ & $103.7_{0.0}$ & $10.2_{0.0}$ & $39.5_{0.0}$ \\
        \hline
        Original LText & $178.1_{12.0}$ & $13.1_{0.4}$ & $34.1_{1.0}$ & $14.2_{0.5}$ & $33.8_{1.3}$ & $100.5_{3.7}$ & $30.0_{1.0}$ & $9.2_{0.1}$ & $40.7_{0.5}$ \\
        Original LVision & $142.3_{10.3}$ & $11.7_{0.1}$ & $36.8_{1.4}$ & $9.7_{0.2}$ & $45.6_{1.7}$ & $105.1_{3.6}$ & $24.4_{0.4}$ & $7.5_{0.1}$ & $40.8_{1.1}$ \\
        Original QText & $284.2_{9.0}$ & $27.0_{0.6}$ & $38.1_{1.2}$ & $28.5_{0.6}$ & $55.1_{0.9}$ & $122.3_{21.6}$ & $28.2_{1.4}$ & $18.0_{0.6}$ & $68.3_{0.4}$ \\
        Original QVision & $270.7_{15.4}$ & $14.3_{0.3}$ & $39.5_{0.5}$ & $8.2_{0.3}$ & $40.2_{2.0}$ & $154.5_{13.0}$ & $42.4_{0.5}$ & $11.2_{0.2}$ & $53.7_{0.5}$ \\
        \hline
        Fine-tuned LText & $\mathbf{99.8_{22.5}}$ & $2.6_{0.1}$ & $27.3_{2.1}$ & $3.8_{0.1}$ & $15.9_{0.5}$ & $105.1_{6.8}$ & $22.8_{0.9}$ & $9.1_{0.1}$ & $20.5_{0.8}$ \\
        Fine-tuned LVision & $119.5_{14.9}$ & $2.6_{0.0}$ & $14.4_{0.7}$ & $\mathbf{3.2_{0.1}}$ & ${12.0_{0.2}}$ & ${78.5_{8.4}}$ & $22.0_{0.4}$ & $5.9_{0.2}$ & $22.6_{0.4}$ \\
        Fine-tuned QText & $101.3_{6.8}$ & $\mathbf{2.3_{0.0}}$ & $\mathbf{13.6_{0.4}}$ & $5.6_{0.3}$ & $11.9_{0.3}$ & $103.1_{8.0}$ & $22.9_{0.1}$ & $\mathbf{5.8_{0.1}}$ & $20.9_{0.4}$ \\
        Fine-tuned QVision & $174.5_{12.5}$ & $2.7_{0.0}$ & $16.2_{0.4}$ & $3.6_{0.1}$ & $14.2_{0.5}$ & $157.9_{8.9}$ & $21.8_{0.9}$ & $6.8_{0.1}$ & $21.1_{0.4}$ \\
        \hline
    \end{tabular}
    \label{tab:all_mape_results}
\end{table*}

We evaluate the models on the Kaggle polymer dataset across five target properties (Tg, FFV, Tc, density, and Rg) and report MAE, MAPE, and wMAE. Following our protocol, VLM results are averaged over five inference runs, while ML baselines are averaged over five independently trained instances. Results for MAE and wMAE are presented in Table~\ref{tab:kaggle_mae_results}. For external validation, we test the models on the GTT and RadonPy datasets. The corresponding MAE and wMAE are shown in Table~\ref{tab:unseen_mae_results}. Additionally, all MAPE results are provided in Table~\ref{tab:all_mape_results}. Note that for the RadonPy dataset, wMAE is computed only over Tc, Density, and Rg, since Tg is evaluated separately on the GTT dataset.

\subsubsection{Performance on Kaggle Test Data} In Table~\ref{tab:kaggle_mae_results}, VLMs demonstrate clear advantages prior to fine-tuning: Original LVision achieves a wMAE of 0.130 and Original QVision 0.136, both outperforming their text-only counterparts, Original LText (0.133) and Original QText (0.220). After fine-tuning, Fine-tuned LVision delivers the best overall performance on the Kaggle benchmark, attaining the lowest wMAE of 0.045 and the best MAE for Density (0.033). Fine-tuned LText and QText perform competitively but still trail LVision, underscoring the importance of visual information in this study. ML baselines trained on RDKit descriptors and PolyBERT embeddings achieve strong results for some targets, but they remain inferior to Fine-tuned LVision in terms of overall wMAE. Specifically, Fine-tuned LVision outperforms Fine-tuned LText (0.054), Fine-tuned QText (0.049), Fine-tuned QVision (0.052), as well as all descriptor- and embedding-based baselines.

In Table~\ref{tab:all_mape_results}, with respect to MAPE, original VLMs already outperform text-only counterparts on Tg, FFV, and Density. For example, Original LVision achieves lower MAPE than Original LText on Tg (142.3\% vs. 178.1\%), FFV (4.5\% vs. 5.2\%), and Density (23.0\% vs. 25.1\%). Similarly, Original QVision performs better than Original QText on Tg (270.7\% vs. 284.2\%), FFV (7.3\% vs. 7.7\%), and Density (28.2\% vs. 28.5\%). After fine-tuning, nearly all fine-tuned models outperform descriptor- and embedding-based baselines in MAPE, except that SVR + PolyBERT achieves the best Rg MAPE at 9.9\%. These results highlight that multimodal fine-tuning surpasses baseline methods on Tg, Tc, FFV, and Density. Among fine-tuned models, Fine-tuned LText achieves the best Tg MAPE (99.8\%), while Fine-tuned QText achieves the lowest errors for FFV (2.3\%) and Tc (13.6\%). In contrast, Fine-tuned LVision attains the lowest Density error (3.2\%).

\subsubsection{Performance on Unseen Datasets}
We evaluate how models fine-tuned on the Kaggle dataset generalize to unseen data. As shown in Table~\ref{tab:unseen_mae_results}, before fine-tuning, Original LVision (0.617) and Original QVision (0.870) already outperform their text-only counterparts in terms of wMAE on the GTT dataset, compared to Original LText and Original QText. After fine-tuning, LVision (0.677) and QVision (0.659) further improve upon LText (0.707) and QText (0.814), highlighting the benefit of multimodal inputs. On the RadonPy dataset, Original LVision achieves a wMAE of 0.617 and Original QVision 0.870, again outperforming LText and QText. More specifically, Original LVision achieves lower MAEs than both LText and QText on Tc (0.059 vs. 0.080 and 0.074), Density (0.086 vs. 0.105 and 0.194), and Rg (10.0 vs. 10.4 and 16.9), demonstrating the consistent value of visual information prior to fine-tuning. After fine-tuning, LVision achieves the best Density MAE (0.071), while QVision records the second-lowest wMAE (0.415), outperforming LText (0.450) and slightly improving upon QText (0.420). Although SVR + PolyBERT achieves the lowest overall wMAE (0.406) and delivers the best results on Tc and Rg, multimodal fine-tuning remains superior for Density prediction and highlights the added value of incorporating visual information.

In Table~\ref{tab:all_mape_results}, original VLMs outperform text-only models primarily on Tc and Density. In particular, Original LVision achieves lower MAPE than both Original LText and Original QText on Tc (24.4\% vs. 30.0\% and 28.2\%) and Density (7.5\% vs. 9.2\% and 11.2\%). After fine-tuning, LVision records the second-lowest Tg MAPE (78.5\%), slightly above MLP + RDKit (75.4\%). For Density, Fine-tuned QText and LVision achieve the lowest and second-lowest MAPE values (5.8\% and 5.9\%, respectively). LVision and QVision also slightly outperform LText and QText on Tc, with MAPEs of  $\sim$22\% compared to $\sim$23\%. As with MAE, descriptor- and embedding-based baselines remain competitive: MLP + RDKit leads on Tg (75.4\%), SVR-based methods excel on Tc (17.7\%), and RF + PolyBERT performs best on Rg (16.7\%). Overall, these results suggest that while traditional baselines remain strong on Tg, Tc, and Rg, fine-tuned vision-based models consistently excel on Density prediction, reaffirming the unique contribution of visual information.

The improved performance of Fine-tuned LVision and QVision is both promising and consistent with prior findings. Previous work has shown that deep learning on 2D molecular depictions can match or surpass string-based representations such as SMILES by enabling augmentation, transfer learning, and visual feature extraction \cite{goh2017chemception, wilkinson2022images}. While a 2D depiction generated by RDKit from a SMILES string does not inherently encode more chemical information than the SMILES itself \cite{o2012towards, krenn2022selfies}, it provides a visually structured representation that enhances how Vision–Language Models process and interpret molecular structures.

\section{Conclusion}
We present a multimodal instruction-tuning dataset for polymer property prediction and fine-tune a Vision–Language Model using images, P-SMILES, and molecular descriptors. We compare its performance to baselines, including traditional ML with RDKit or PolyBERT features, and Llama-based vision and text-only models. Our fine-tuned Llama-3.2-11B-Vision-Instruct achieves competitive performance, highlighting the value of multimodal inputs. Moreover, it removes the need for separate per-property models, thereby reducing deployment and maintenance costs. In future work, we will expand the dataset to include a larger number of samples. Incorporating additional molecular descriptors may further improve VLM performance and help them surpass baseline models. Finally, it will be important to explore how to build a foundation model that integrates additional modalities, such as 3D conformations or graph-based representations of polymers. Developing foundation models for new polymer discovery represents a critical challenge and an exciting opportunity for advancing AI in materials science.

\section*{Acknowledgment}
This work was supported in part by the National Institute of General Medical Sciences of National Institutes of Health under award P20GM139768, and the Arkansas Integrative Metabolic Research Center at the University of Arkansas.

\bibliographystyle{ieeetr}
\bibliography{references.bib}

\end{document}